\documentclass[10pt,twocolumn,letterpaper]{article}

\usepackage{cvpr}              
\usepackage{amsmath,graphicx}
\usepackage[utf8]{inputenc} 
\usepackage[T1]{fontenc}    
\usepackage{natbib}
\setcitestyle{numbers,square}
\usepackage{url}            
\usepackage{multirow}

\usepackage{amssymb}
\usepackage{algorithm}
\usepackage{xparse}
\usepackage{algpseudocode}
\usepackage{bm}
\usepackage[dvipsnames]{xcolor}

\definecolor{cvprblue}{rgb}{0.21,0.49,0.74}
\usepackage[pagebackref,breaklinks,colorlinks,citecolor=cvprblue]{hyperref}

\newcommand{\tabincell}[2]{\begin{tabular}{@{}#1@{}}#2\end{tabular}}

\def\paperID{12271} 
\def\confName{CVPR}
\def\confYear{2024}

\title{Pursing the Sparse Limitation of Spiking Deep Learning Structures}

\author{Hao Cheng\\
HKUS(GZ)
\and
Jiahang Cao\\
HKUST(GZ)
\and
Erjia Xiao\\
HKUST(GZ)
\and
Mengshu Sun\\
Beijing University of Technology
\and
Le Yang \\
Xi'an Jiaotong University
\and
Jize Zhang \\
HKUST
\and
Xue Lin \\
Northeastern University
\and
Bhavya Kailkhura \\
Lawrence Livermore National Laboratory 
\and
Kaidi Xu \\
Drexel University 
\and
Renjing Xu\\
HKUST(GZ)
}

\begin{document}
\maketitle
\begin{abstract}

Spiking Neural Networks (SNNs), a novel brain-inspired algorithm, are garnering increased attention for their superior computation and energy efficiency over traditional artificial neural networks (ANNs). To facilitate deployment on memory-constrained devices, numerous studies have explored SNN pruning. However, these efforts are hindered by challenges such as scalability challenges in more complex architectures and accuracy degradation.
Amidst these challenges, the Lottery Ticket Hypothesis (LTH) emerges as a promising pruning strategy. It posits that within dense neural networks, there exist winning tickets or subnetworks that are sparser but do not compromise performance. 
To explore a more structure-sparse and energy-saving model, we investigate the unique synergy of SNNs with LTH and design two novel spiking winning tickets to push the boundaries of sparsity within SNNs.
Furthermore, we introduce an innovative algorithm capable of simultaneously identifying both weight and patch-level winning tickets, enabling the achievement of sparser structures without compromising on the final model's performance.
Through comprehensive experiments on both RGB-based and event-based datasets, we demonstrate that our spiking lottery ticket achieves comparable or superior performance even when the model structure is extremely sparse.

\end{abstract}    
\section{Introduction}
\label{sec:intro}

SNN is acclaimed as the third generation of neural networks and has increasingly gained great interest from researchers in recent years due to its distinctive properties: high biological plausibility, temporal information processing capability, inherent binary (spiking) information processing superiority, and low power consumption.
Unlike ANNs which represent data continuously, SNNs process information as binary time series, leveraging low-power accumulation (AC) operations instead of the power-intensive multiply-accumulate (MAC) operations common in ANNs. This fundamental difference not only enhances energy efficiency but also aligns closely with biological neural processing. 
On specialized neuromorphic hardware platforms such as Loihi~\cite{davies2018loihi} and TrueNorth~\cite{akopyan2015truenorth}, SNNs demonstrate a remarkable reduction in energy consumption compared to ANNs. 
Additionally, SNNs follow their biological counterparts and inherit complex temporal dynamics from them, endowing SNNs with powerful abilities to extract image features in a variety of tasks, including recognition~\cite{zhou2022spikformer,deng2022temporal}, tracking~\cite{zhang2022spiking}, and images generation~\cite{cao2023spiking}.

\begin{figure*}[t]
	\setlength{\tabcolsep}{1.0pt}
	\centering
	\begin{tabular}{c}
		
		\includegraphics[width=1.0\textwidth]{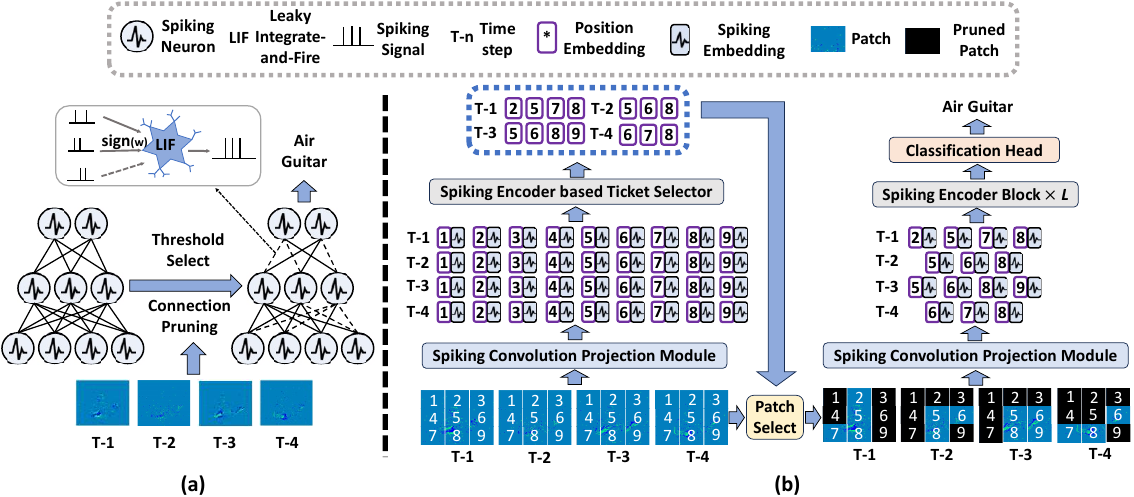} \\
		
	\end{tabular}
	\caption{ The Spiking Lottery Tickets (SLT) finding in event-based data (a) the finding process in spiking-based CNNs; (b) the finding process in the Spiking-based Transformers. }
	\label{fig:all_structure}
	\vspace{-0.35cm}
\end{figure*}

Despite the advancements and comparable performance of recent SNN algorithms to ANNs in various domains, practical challenges such as operational redundancy and computational load persist, hindering their broader adoption in hardware implementations. To mitigate these issues, recent studies have focused on pruning techniques in SNNs, exploring methods such as model pruning~\cite{neftci2016stochastic,rathi2018stdp,chen2021pruning}, model quantization~\cite{xu2020generative, zhou2018adaptive, fang2021deep}, and knowledge distillation~\cite{polino2018model, chen2022state}. 
However, these methods face their own set of limitations. Many are constrained to simpler models~\cite{neftci2016stochastic,rathi2018stdp}, or they lead to significant performance degradation~\cite{chen2022state, fang2021deep}, especially in more complex spiking model architectures. This gap underscores the need for more effective pruning strategies that can maintain or enhance performance while accommodating the intricate dynamics of advanced SNNs.

Confronting the aforementioned challenges, the Lottery Ticket Hypothesis (LTH) emerges as a promising avenue for pruning in SNNs. LTH posits that within a randomly initialized, dense neural network, there exists a subnetwork, which trained independently, can achieve the test accuracy of the full network within the same or fewer iterations. 
This approach to sparse training contrasts with the conventional multi-step process of weight pruning involving pretraining, pruning, and fine-tuning, and allows for the discovery of these efficient subnetworks through a single phase of weight training. Building upon this concept, the Multi-Prize Lottery Tickets (MPTs~\cite{diffenderfer2021multiprize}) hypothesis further refines this approach by focusing on efficient connection selection without the necessity of weight training, enhancing both weight sparseness and binarization for improved performance. Inspired by these advancements, our study aims to delve into the synergistic potential between LTH and the inherent energy efficiency of SNNs. We seek to construct models that are not only structurally sparse but also excel in energy conservation, thereby pushing the boundaries of SNN capabilities with the LTH method.


However, existing LTH works have been primarily tailored for widely used Convolutional Neural Networks (CNNs) and Vision Transformers (ViT). Moreover, integrating LTH with SNNs is relatively less explored and still problematic. 
For instance, Kim et al.~\cite{kim2022exploring} proposed an early-time (ET) ticket for deep SNNs, a notable advance but still reliant on extensive iterative processes to uncover winning tickets. Yao et al.~\cite{yao2023probabilistic} introduced a probabilistic modeling approach for LTH in SNNs, yet its validation is limited to a narrow dataset range and standard CNN structures, potentially underutilizing SNNs' distinct capabilities.



In response to the gaps identified in existing studies, our research endeavors to delve into the unique synergy between Spiking Neural Networks (SNNs) and the Lottery Ticket Hypothesis (LTH). Our investigation is driven by two pivotal questions pivotal to advancing the concept of spiking lottery tickets (SLT): 
(1) Existence of SLTs in Event-Based Data. SNNs are commonly adopted for handling event-based data since utilizing SNNs on neuromorphic hardware is low-energy and low-latency~\cite{roy2023live,viale2021carsnn}. Our work aims to uncover and characterize SLTs across various SNN configurations in event-based data scenarios, shedding light on their potential and limitations.
(2) Winning Tickets in Spiking-Based Transformers: The emergence of spiking-based transformers~\cite{zhou2022spikformer,wang2023masked} has opened new frontiers in neural network architectures, showing promise for future deployment in edge devices and real-world applications. However, the exploration of patch-winning tickets within these spiking-based transformers remains largely unexplored. Our study seeks to fill this research void, examining whether these advanced models harbor winning tickets that could further enhance their efficiency and applicability.

To address the challenges previously outlined and to conduct a thorough exploration of Spiking Lottery Tickets (SLTs), our contributions can be summarized as follows:
\begin{itemize}
        \item \textbf{From RGB to DVS:} For the lottery tickets hypothesis, we remedy its consideration under event-based datasets. The DVS128Gesture~\cite{amir2017low} and CIFAR10-DVS~\cite{li2017cifar10} are adopted to explore the event-based winning tickets.    
	\item  \textbf{From CNN to ViT:} 
 Building upon the discovery that weight-level SLTs are present in spike-based CNNs across both RGB and event-based data, we broaden our investigation to spiking-based transformers. Through our extensive real-world explorations, it can be shown that patch-level SLTs also exist in different kinds of datasets.
 	\item  \textbf{SNNs Parameters Analysis:} Having fully explored the winning tickets discovery algorithm's impact on SLTs, we investigated the effect of the intrinsic parameters of the SNNs themselves on SLTs.
        \item  \textbf{Concurrent Connection and Token level SLT:} After a comprehensive analysis of the characteristics of SLTs for both weight and patch level, we propose a novel algorithm to implement both SLTs in spiking-based transformers.        
\end{itemize}

\section{Related Works}
\subsection{Model Structures}

Within the realm of deep learning, Deep Neural Networks (CNNs) have emerged as a fundamental architecture for many computer vision tasks, driving forward the state-of-the-art with their capacity to learn complex hierarchical visual features.

\noindent\textbf{Artificial Neuron Networks}
Among various structures, Convolutional Neural Networks~\cite{lecun1998gradient} became the cornerstone and profoundly influenced all recent related research. After that, the evolution of CNNs are proposed, like VGG~\cite{simonyan2014very}, ResNet~\cite{he2016deep}, DenseNet~\cite{huang2017densely}, and others. These architectures have excelled in image classification tasks and have been foundational in advancements across various applications in computer vision, such as object detection, semantic segmentation, and more. Their widespread popularity is a testament to the versatility and robustness of CNNs as a class of deep learning models. Following the enormous success of CNNs, a novel structure Vision Transformer (ViT), which is the encoder part of the Transformer, has received a great deal of attention. ViTs apply the transformer architecture, traditionally used in natural language processing, to vision tasks, treating images as sequences of pixels or patches. This seminal work demonstrated that transformers could outperform CNNs on image classification tasks when trained on sufficiently large datasets and with enough computational resources. Following the initial ViT model, various adaptations have emerged to enhance performance and efficiency. For instance, the DeiT (Data-efficient image Transformers) modified the training procedure to make the transformer models more data-efficient, expanding their applicability to scenarios where large-scale labeled datasets are not available. Incorporating convolutional information into the original Vision Transformer (ViT) architecture has been a subject of interest to improve its performance, especially in terms of inductive biases and local feature processing. Below are a few architectures that have successfully integrated convolutional elements into ViTs~\cite{dosovitskiy2020image}. Levit~\cite{graham2021levit}, LV-VIT~\cite{jiang2021all}, CVT~\cite{wu2021cvt}, Swin~\cite{liu2021swin} and others~\cite{cheng2023rbformer}.

\noindent\textbf{Spiking Neuron Networks}
This section is about the current research progress of spiking neuron networks.
The spiking neural network is a bio-inspired algorithm that simulates the real process of signaling that occurs in brains. Compared to the artificial neural network (ANN), it transmits sparse spikes instead of continuous representations, which brings advantages such as low energy consumption and robustness. 
In this paper, we adopt the widely used Leaky Integrate-and-Fire (LIF) model~\cite{roy2019towards}, which is suitable to characterize the dynamic process of spike generation and can be defined as:
\begin{equation}
    \tau\frac{\mathrm{d} V(t)}{\mathrm{d} t}= - (V(t) - V_{reset}) + I(t)
\end{equation}
where $I(t)$ represents the input synaptic current at time $t$ to charge up to produce a membrane potential 
  $V(t)$, $\tau$ is the time constant. When the membrane potential exceeds the threshold $V_{th}$, the neuron will trigger a spike and resets its membrane potential to a value $V_{reset}$ ($V_{reset}<V_{th}$). The LIF neuron achieves a balance between computing cost and biological plausibility. 

In practice, the dynamics need to be discretized to facilitate reasoning and training. The discretized version of LIF model can be described as:
\begin{align}
    & U[n] = e^{\frac{1}{\tau}}V[n-1] + (1-e^{\frac{1}{\tau}}) I[n] \label{eq:dis_lif1}\\
    & S[n] = \Theta (U[n] - V_{th})\label{eq:dis_lif2}\\
    & V[n] = U[n](1-S[n]) + V_{reset}S[n] \label{eq:dis_lif3}
\end{align}
where $n$ is the discrete time step, $U[n]$ is the membrane potential before reset, $S[n]$ denotes the output spike which equals 1 when there is a spike and 0 otherwise, $\Theta(x)$ is the Heaviside step function, $V[n]$ represents the membrane potential after triggering a spike. 

\subsection{Model Sparsity}
The model sparsity is to pursue the model with fewer model parameters without sacrificing the final performance.

\noindent\textbf{Artificial Neuron Networks Sparsity}
Deep learning compression pursues more lightweight model parameters that could facilitate DNN implementations on resource-constrained application systems. 
Among them, model pruning and quantization are my primary focus that could be more possibly applied to hardware applications, such as FPGA.
About model pruning, there is in general regular pruning
schemes that can preserve the model's structure in some sense, such as the filter pruning scheme~\cite{luo2017thinet,he2019filter}, block-based pruning scheme~\cite{dong2020rtmobile,ma2020blk}, and otherwise irregular pruning scheme~\cite{he2017channel,wen2016learning,zhang2018adam}. 
After the success of those different kinds of weights training, 
The lottery ticket hypothesis (LTH) \cite{frankle2018lottery} conjectures that inside the large network, a subnetwork together with their initialization makes the pruning particularly effective, and together they are termed as the ``winning tickets". In this hypothesis, the original initialization of the sub-network (before the large network pruning) is significant for it to achieve competitive performance when trained in isolation. Additionally, LTH also emphasizes the significance of the submodel structure itself but not the inheriting weights or gradients of the initial pre-trained models. 
This hypothesis also brings the possibility of Sparse Training~\cite{ding2021optimal, diffenderfer2021multiprize, liu2020dynamic} that could
realize the finding of sparse subnetworks without complex collocation techniques of pertaining, pruning, and retraining cyclically. 
Model quantization~\cite{zhou2018adaptive, chmiel2020robust, tang2017train, yang2017bmxnet, lin2020rotated} is another strategy to facilitate model compression. It could quantize the original 32-bit model to 16-2 bit models. Among them, converting 32 bits to 2 bits is the extreme situation of quantization that could be termed the Binary Neural Network (BNN)~\cite{tang2017train, yang2017bmxnet, lin2020rotated}. Model binarization could transfer original mathematics computation to XNOR or other bit-wise operations that could deeply boost high-performance computing in different devices. Additionally, BNN is also more suitable for the application of various hardware since the basis of digital circuits is the binary representation. MPTs~\cite{diffenderfer2021multiprize} as the extension of LTH~\cite{diffenderfer2021multiprize} is adopted to certify that the lottery tickets or subnetworks also remain in quantization or binary dense models.

\noindent\textbf{Spiking Neural Networks Sparsity}
To further improve the energy efficiency of SNN, a number of works on SNN pruning have been proposed and well-validated on neuromorphic hardware. Shi~\cite{shi2019soft} propose a pruning scheme that exploits the output spike firing of the SNN to reduce the number of weight updates during network training. Guo~\cite{guo2020unsupervised} dynamically removes
non-critical weights in training by using the adaptive online pruning algorithm. Apart from seeking the help of pruning, Rathi~\cite{rathi2018stdp} and Takuya~\cite{takuya2021training} pursue sparse SNN by using Knowledge distillation and quantization. 
Several works try to combine LTH with SNNs: Kim et al.~\cite{kim2022exploring} first investigate how to scale up pruning techniques towards deep SNNs and reveal that winning tickets consistently exist in deep SNNs across various datasets and architectures. They also propose a kind of Early-Time ticket that could alleviate the heavy search cost. Yao et al.~\cite{yao2023probabilistic} contribute a novel approach by introducing a probabilistic modeling method for SNNs. This method allows for the theoretical prediction of the probability of identical behavior between two SNNs, accounting for the complex spatio-temporal dynamics inherent to SNNs. Cheng et al.~\cite{cheng2023gaining} refers to \textit{biprop}~\cite{diffenderfer2021multiprize} and extends the discovered scenario to binary weight case thereby further exploring the sparse limit of spiking winning tickets. Concretely, they compare the spiking mechanism with simple model binarization and propose a superior strategy for finding SLT named Binary Weights Spiking Lottery Tickets. The binary SLT explored by this paper could even perform better than the original dense ANN and BNN in some cases.





\noindent\textbf{Training Spiking Neural Network}
In the past few years, a large number of learning algorithms have explored how to train a deep SNN with high performance, including three main approaches: (1) ANN to SNN conversion, (2) direct training, and (3) local training. For (1), it is necessary to train an ANN network in advance and then transfer the continuous output of the ReLU function to the discrete spikes~\cite{ding2021optimal,bu2021optimal}. Although such algorithms can help SNNs achieve better accuracy, they require substantial training resources. The second type of method usually employs a gradient-based technique. For example, the surrogate gradient method~\cite{neftci2019surrogate} solves the non-differentiability problem of discrete spikes during the backpropagation process. Meng ~\cite{meng2022training} proposes DSR to train SNN indirectly by gradient mapping. Although such methods ensure the feasibility of training, they do not exploit the properties of SNNs and lack bio-plasticity. Lastly, the local training methods are mainly based on biological learning rules, like Hebbian learning~\cite{hebb2005organization} and Spike Timing-Dependent Plasticity (STDP)~\cite{qi2018jointly,rathi2018stdp}. Although the performance of those methods is much lower than the State-of-the-art (SOTA) ANN, the local training is particularly sparse, which provides ideas for SNN model compression. 
In general, some methods are inspired by existing compression techniques in ANN. 
Kim~\cite{kim2022exploring} proposes the Early-Time ticket to relatively reduce the huge training computational cost when combined with the multiple timesteps of SNNs.



\section{Spiking Lottery Tickets}
To go about enhancing the generalizability of previous SLTs explorations, 
Regarding Ramanu et al.~\cite{ramanujan2020s} and Shen et al.~\cite{shendata} that are the discovery algorithms towards connection and path-based winning tickets in spiking-based CNNs and Transformers, as well as some the work about SNNs sparsity~\cite{yao2023probabilistic, cheng2023gaining}, we first try also to pursue the extensions of the universality of SLTs in different dataset and structure, like event-based data and spiking-based transformer.
After that, since the spiking-based Transformer or SpikeFormer always needs the convolution operation in embedding to promote its performance, it would be possible to concurrently adopt the former two SLTs finding algorithms to boost the sparsity of Spikeformer further. Therefore, we propose an algorithm that could efficiently find the ECPTs.

\noindent\textbf{Connection-Level Lottery Tickets for CNNs:}
Referring to the main idea of Ramanu et al.~\cite{ramanujan2020s}, they propose the \textit{edge-popup} that could explore the winning tickets in randomly weighted neural networks, which are subnetworks that can achieve high accuracy without any traditional weights training. It shows that identifying high-performing subnetworks does not need to adjust the original weights or rely on the trained weights. In a word, \textit{edge-popup} mainly focuses on finding the Sparse
Connection-based SLTs (SConnSLT) and does not care about the effect of weight values.
Furthermore, Diffenderfer et al. ~\cite{diffenderfer2021multiprize} also refer to \textit{edge-popup} and broaden its applying domain in binary case. In the spiking research area, based on it, Yao et al.~\cite{yao2023probabilistic} and Cheng et al.~\cite{cheng2023gaining} apply those subnetworks finding methods to convolutional-based SNNs and could also efficiently find SLTs without training. 
Fig.~\ref{fig:all_structure} (a) briefly illustrates the process of finding SConnSLT under event-based datasets. Since the time step of event-based data is somewhat similar to the frames in RGB data, it could be visualized as four frames when the Timestep = 4. After inputting the event data into the convolution-based SNNs, the neuron connection would be gradually sparse after the Threshold Select and Connection Pruning. Additionally, the information adopted in the whole process is the spiking signal as presented.

\noindent\textbf{Patch-Level Lottery Tickets for Transformer:}
Shen et al.~\cite{shendata} explore extending the winning tickets to ViTs by focusing on input data, specifically image patches, instead of network weights.
It introduces a Ticket Selector to identify patch-level winning tickets that can train ViTs to achieve accuracy comparable to using all patches. 
Unlike traditional weight-level winning tickets, they focus on data-level 'winning tickets' rather than weight-level due to the unique challenges of ViTs. 
ViTs heavily rely on input data, making it difficult to generalize winning subnetworks across different inputs using weight-level approaches. Traditional methods of finding effective subnetworks at the weight level are less effective in ViTs, as their performance is more tied to input handling. By identifying the most informative image patches, this approach aims to train ViTs more efficiently and effectively, shifting the focus of the Lottery Ticket Hypothesis from network weights to input data.
To the best of our knowledge, no article has attempted to explore any form of spiking winning tickets for spiking ViTs, regardless of whether the SLTs are weight or data-based, and regardless of whether the SLTs are found in RGB data or event data.
As the illustration of Fig.~\ref{fig:all_structure} (b), it indicates the process of finding date-based SLTs in event-based data.
Specifically in ViTs, the data-based Winning Tickets actually screen for the importance of different patches and tokens.
The left subfigure of Fig.~\ref{fig:all_structure} (b) shows the Token-based Lottery Tickets (TokenLTs) could be identified after going through the Spiking Patch Splitting Module (SPS) and Spiking Encoder-based Ticket Selector (SETS). After obtaining the TokenLTs and their position embedding through the patch selection to the original patch, the insignificant patches are deleted and only the TokenLTs go through the concrete classification process in the right subfigure of Fig.~\ref{fig:all_structure} (b). 

\begin{algorithm}[h!]
\caption{Embedding Connection and Patch Tickets}
\begin{algorithmic}[1]
\State \textbf{Define Models:} Spikeformer $S(\cdot)$; spiking Convolution Projection Module (CPM) $S_{c}$ with param numbers $n_{c}$; Spiking Encoder and classification head $S_{ec}(\cdot)$.
\State \textbf{Input:} Event-based data  $(x_{t}, y)$ with patch $P$ and time $t$; Patch number $n_p$; Connection tickets finding epochs $N_{c}$; Spiking Patch Tickets (SPT) Select epochs $N_{sp}$.   
\State \textit{Initialize Model Params:} The CPM weights and its corresponding sparse mask $w_{c}$ and $M_{c} \in \{0, 1\}$;
\State \#\#\# Connection lottery tickets finding in CPM
\State \textit{Initialize Params:} Pruning score $s$ and its optimizing update param $\eta$; Gain term $\alpha$; Pruning rate for connection sparsity $pr_c$; CPM loss function $L(\cdot)$
\State \textit{Adding a linear layer with the output dimension $y$}. 
\For{k = 1 to $N_{c}$}
\State $s \gets s -  \eta \nabla_{s} L(\alpha \cdot M_c \odot w_{c})$
\State $Proj_{[0,1]} \gets $ Sorting  $s$ based on to $pr_{c} n_c$
\State $M_c \gets M_c \odot Proj_{[0,1]}$, $\alpha \gets  ||M_c \odot w_c|| / ||M_c|| $ 
\EndFor
\State \textit{Keep the convolution layer and drop the linear layer}.
\State \#\#\# Patch lottery ticket finding for spikeformer
\State \textit{Initialize Params:} SPT index $id_p$;
Patch embedding and Position embedding $PatE$ and $PosE$; pruning rate for patch sparsity $pr_p$. 
\For{k = 1 to $N_{cp}$}
\State $ PatE \gets S_{c} (P)$
\State $id_p \gets $ Sorting  $PatE$ based on $pr_{p} n_{p} \odot PosE$
\EndFor
\State  $P_{spt} \gets$ Pick the SPT according to the $id_p\odot P$
\State \textbf{Output}: $S(p_{spt}) \gets S_{ec}(S_{p}(P_{spt}; \alpha(M_c \odot w_{c})))$
\end{algorithmic}
\label{Al:comb}
\end{algorithm}

\noindent\textbf{Concurrent Connection and Patch Level Lottery Tickets:}
As presented in Fig.~\ref{fig:all_structure}, the previous two subsections propose the lottery ticket finding strategies for CNN and transformer-based SNNs on connection and patch level respectively. However, when we revisit the supposedly lightweight spikeformer later on, it seems that there is still the possibility of improving sparsity even more. ViTs are less likely to get winning tickets at the connection and weights level, because they consist of only a few linear layer-based attention operations, and there are no convolution operations that might cause redundancy. Nonetheless, nowadays, the introduction of convolution operation into ViTs has become a mainstream performance improving strategy. Furthermore, in the spiking domain, this operation seems to have become an even more necessary component to improve the performance of the corresponding structures, e.g., the input embedding of the spikeformer adopts the convolution projection to process input patch directly. This provides us with the feasibility to further enhance the connection level sparse of patch-level tickets of Spikeformer. Therefore, by combining the ideas of Figure~\ref{fig:all_structure}, we propose the Embedding Connection and Patch Tickets (ECPTs) finding algorithm for Transformer-based SNNs. 

\begin{figure}[h]
	\setlength{\tabcolsep}{1.0pt}
	\centering
	\begin{tabular}{c}
		
		\includegraphics[width=0.45\textwidth]{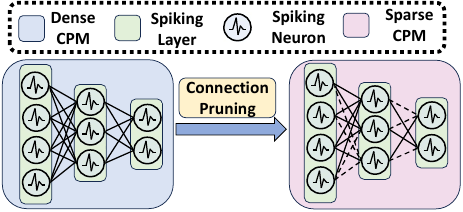} \\
	\end{tabular}
	\caption{Connection sparsity in Convolution Projection Module}
	\label{fig:cit_cnn_lth}
	\vspace{-0.35cm}
 \label{fig:2}
\end{figure}

Since the purpose of finding ECPTs is mainly to improve further the connection-level winning tickets in the spiking Convolution Projection Module (CPM) of the previous patch-level tickets in Spikeformer, the significant modification of the original finding method in the subfigure (b) in Figure ~\ref{fig:all_structure} should be the introduction of connection level tickets finding process towards CPM. As illustrated in Figure~\ref{fig:cit_cnn_lth}, after applying the connection pruning to the original CPM, some redundancy connection would be pruned according to the particular pruning rate, and the initial dense CPM is transferred to the sparse CPM.  

The Algorithm~\ref{Al:comb} presents further details of the discovery of ECPTs. Step 1 is adopted to define the model. The Spikeformer adopted here is $S(\cdot)$, and our mainly sparse object in finding ECPTs is in the spiking CPM. After that, the remaining structures, which are the spiking encoder block and classification head, of Spikeformer is $S_{ec}$. And Steps 2 and 3 are the input parameters and the initialization of the part of them.  And then,  step 4 to step 11 is the main connection ticket finding process. The main parameter of this finding process is the pruning score $s$ and it could be continuously updated until finding the correct connection tickets according to the conjunction of mask $M_c$, weights $w_c$, and gain term $\alpha$.  However, this gain term update is driven by optimizing the parameters of a to achieve the search for subnetworks with optimal connections in a continuous update. After finding the corresponding connection tickets, the updated mask $M_c$ and weights $w_c$ would be kept to insert into the original Spikeformer. Consequently, step 15 to step 19 would finsh the patch tickets finding phase. The original patches $P$ are fed into CPM $S_c$ and output patch embedding $PatE$. The index $id_p$ of the most significant embedding or patch tickets would be selected according to the sorting of $PatE$.  Finally, the picked patch winning tickets $P_{spt}$ would be processed in this Spikeformer with the sparse CPM.

\section{Experiments and Analysis}

\begin{table*}[h]
    \centering
    {
    \small 
    \setlength\tabcolsep{8pt}
    \begin{tabular}{c|c|cccc}
    \toprule[1.2pt]
     \multicolumn{1}{c|}{\bf Architecture} & \multicolumn{1}{c|}{\bf Method}   & \multicolumn{1}{c}{\bf \tabincell{c}{CIFAR10\\ Acc(\%)}} &  \multicolumn{1}{c}{\bf \tabincell{c}{CIFAR100\\ Acc(\%)}}        &  \multicolumn{1}{c}{\bf \tabincell{c}{DVS128Gesture\\ Acc(\%)}} & \multicolumn{1}{c}{\bf \tabincell{c}{CIFAR10-DVS\\ Acc(\%)}} \\
    \toprule[1.0pt] 
    \multirow{10}{1in}{\centering\emph{VGG-9 / ResNet-19}\\(2.26M) / (12.63M)} 
    & ANN$+$  &88.10/92.05  & 68.64/76.74 & 92.92/94.83  &  78.65/80.14\\
    & BNN$*$  & 86.75/91.54 & 67.27/74.42 & 92.09/93.69  & 77.94/78.98\\
    & BinActBNN$^{-}$  & 83.84/86.74 &  58.57/66.60  & 85.41/89.43 &  69.21/62.21\\
    & SNN$*$  & 87.71/91.78 & 67.94/72.37 &  92.78/93.27  & 78.41/80.06\\
    & BinWSNN$-$  & 87.57/91.21 & 66.81/72.93 &  91.97/93.92  & 77.49/79.04\\
    &ANN-ConnS$+$  &87.73/90.87  & 67.56/75.62 & 92.43/93.83  &  77.92/80.02\\
    &BNN-ConnS$*$  & 85.27/91.07 & 66.73/73.87 & 91.74/93.11  & 77.31/78.26\\
    &ActBNN-ConnS$^{-}$  & 81.30/83.96 &  55.81/62.76  & 81.53/84.77 &  63.82/57.41\\
    &SNN-ConnS$^{*}$  & 87.52/91.36 & 67.22/72.01 &  92.17/92.88  & 77.94/79.71\\
    &\textbf{BinWSNN-ConnS}  & \textbf{87.76/91.82} & \textbf{66.74/72.73} &   \textbf{91.70/93.52}  & \textbf{77.01/78.66} \\ 
    \bottomrule[1.2pt]
    
    \end{tabular}
    }\vspace{0.2em}
    \caption{{Evaluating the Connection Sparsity (ConnS) performance of VGG-9/ ResNet-19 in full-dense, binary and spiking condition under CIFAR10, CIFAR100, DVS128Gesture and CIFAR10-DVS. The font bold row \textbf{BinWSNN-ConnS} indicates the model with good performance and the most sparsity among all of them; The $-$ means that \textbf{BinWSNN-ConnS} could greatly over the models of this row; $*$ means that the performance of \textbf{BinWSNN-ConnS} should be almost the same compared with them; $+$ means that \textbf{BinWSNN-ConnS} aims to approach them.
    } 
    }
    \label{tab:cnn}
\end{table*}

\begin{figure*}[h]
	\setlength{\tabcolsep}{1.0pt}
	\centering
	\begin{tabular}{c}
        \includegraphics[width=1\textwidth]{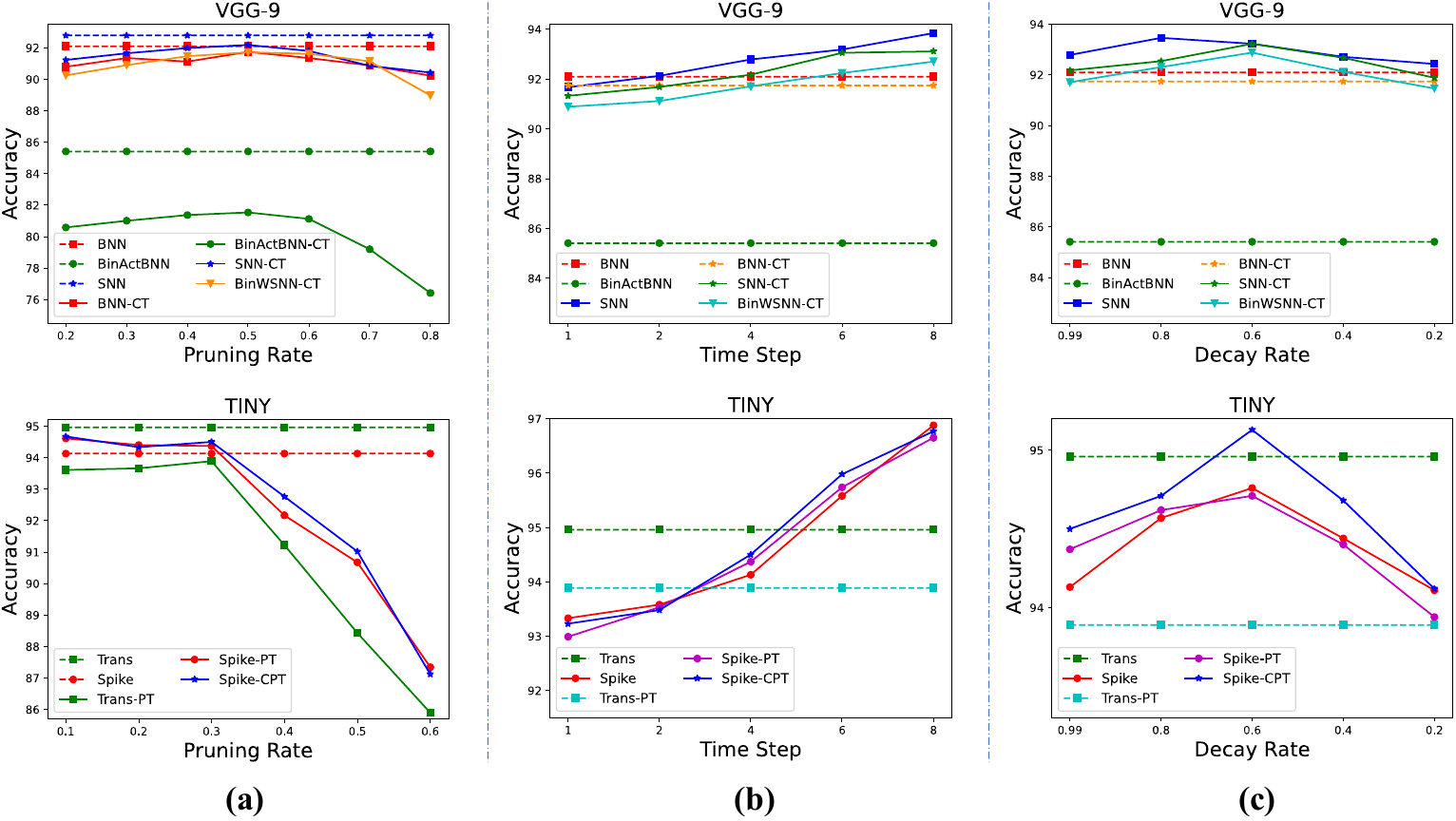}
	\end{tabular}
	\caption{ The performance effect of different model and spiking parameters for the VGG-9 and TINY in DVS128Gesture. The first row indicates the VGG-9, the second row indicates the TINY. The column (a) is the effect of Pruning Rate; The column (b) is the effect of Time Step; The column (c) is the effect of Decay Rate.
        }
	\label{fig:figdecay}
	\vspace{-0.35cm}
 \label{fig:3}
\end{figure*}


\begin{table*}[h]
    \centering
    {
 \small 
    \setlength\tabcolsep{7pt}
\begin{tabular}{c|c|ccccc}
    \toprule[1.2pt]
\textbf{Architecture}                                                                     & \textbf{\begin{tabular}[c]{@{}c@{}}Spasrity \\ and Datasets\end{tabular}} & Transformer$^*$ & Spikeformer$^*$  & \begin{tabular}[c]{@{}c@{}}Transformer\\ - PatchS$^*$ \end{tabular} & \begin{tabular}[c]{@{}c@{}}Spikeformer\\ - PatchP$^*$ \end{tabular} & \textbf{\begin{tabular}[c]{@{}c@{}}Spikeformer\\ - CPS\end{tabular}} \\
    \toprule[1.2pt]
\multirow{6}{*}{\begin{tabular}[c]{@{}c@{}}TINY / SMALL\\ (4.89M) / (22.12M)\end{tabular}} & \textit{Patch Sparsity (\%)}                                                            & 0/0                  & 0/0                  & 31.4/37.6                                                                   & 31.4/37.6                                                                   & \bf 31.4/37.6                                                                \\
& \begin{tabular}[c]{@{}c@{}}\textit{Conn Sparsity(\%)}\end{tabular}           & -/-                  & 0/0                  & -/-                                                                     & 0/0                                                                     & \bf 48.7/54.9                                                                 \\ \cline{2-7} 
& CIFAR10 (\%)                                                                   & 85.71/87.36               & 84.81/87.17               & 82.04/83.76                                                                  & 81.75/82.98                                                                  & \bf 81.21/82.44                                                               \\
& CIFAR100 (\%)                                                                 & 67.37/69.12               & 66.54/69.43               & 64.71/66.22                                                                  & 62.98/64.86                                                                  & \bf 62.82/64.89                                                               \\
& DVS128Gestrure  (\%)                                                          & 94.96/97.12              & 94.13/96.99               & 93.89/96.74                                                                  & 94.37/96.88                                                                  & \bf 94.50/97.22                                                               \\
& CIFAR10-DVS (\%)                                                              & 78.43/80.01               &  77.86/79.77               & 76.61/ 78.42                                                                 & 77.14/78.91                                                                  & \bf 77.03/79.02                    \\                                           
    \bottomrule[1.2pt]
\end{tabular}
    }\vspace{0.2em}
    \caption{Evaluating the Patch Sparsity (PS) and Connection Patch Sparisty (CPS) performance of TINY/SMALL in normal and spiking based Transformer under CIFAR10, CIFAR100, DVS128Gesture and CIFAR10-DVS. The Patch and Connection (Conn) Sparsity are also presented.The font bold row \textbf{Spikeformer-CPS} indicates the model with good peroformance and the most sparsity among all of them;  $*$ means that the performance of \textbf{Spikeformer-CPS} should be almost the same compared with them; }
    \label{tab:vit}
\end{table*}


\begin{figure}[h]
\centering
	\setlength{\tabcolsep}{1.0pt}
	\centering
	\begin{tabular}{c}
        \includegraphics[width=0.4\textwidth]{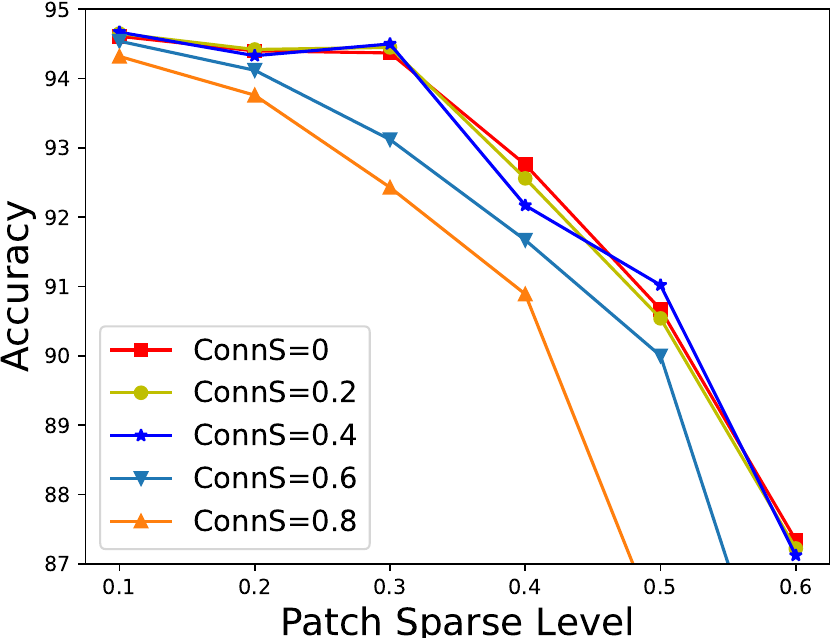}
	\end{tabular}
	\caption{ The performance of patch tickets in spiking-based transformer after introducing Connection Sparse (ConnS) into CPM.
        }
	\label{fig:cpt}
	\vspace{-0.35cm}
 \label{fig:4}
\end{figure}

\subsection{Experimental Setting}
In this paper, we use various strcutres including VGG-9, ResNet-19, TINY and SMALL under two RGB datasets and two Event-based datasets. The TINY and SMALL are referring to the DeiT-Tiny and DeiT-Small~\cite{touvron2021training}. Two RGB datasets are CIFAR10 and CIFAR100. Two Event-based dataset are DVS128Gesture and CIFAR10-DVS. When finding winning tickets, for exploring weight-level tickets using connection sparsity, we adopt $pr_c = 0.2, 0.3, 0.4, 0.5, 0.6, 0.7, 0.8$; for exploring patch level tickets for transformer, we adopt $pr_p = 0.2, 0.3, 0.4, 0.5, 0.6, 0.7, 0.8$.
For the parameters of SNN, we adopt the most commonly used LIF neuron with timestep $T=4$ and decay rate $\lambda=0.99$ in our main experiments. 
Additionally, in order to explore the specific relationship between various SNN component parameters and MPSTs, we 
further use Time Step $T=1,2,4,6,8$ and Decay Rate $\lambda=0.99, 0.8, 0.6, 0.4, 0.2$. To facilitate the learning process, we use the Adam optimizer with a base learning rate of $0.1$. Surrogate gradient training methods~\cite{neftci2019surrogate} for SNN are adopted. The models for conducting experiments are implemented based on Pytorch and SpikingJelly~\cite{SpikingJelly}.
Since our main focus is finding the best spiking lottery tickets while maintaining accuracy, we only use the most primitive direct training without any training tricks in all our experiments. Furthermore, the Appendix includes more detailed interpretations.

\subsection{General Performance Analysis}
According to the results illustrated in Table~\ref{tab:cnn} and Table~\ref{tab:vit}, we could obtain a general performance analysis for the various winning tickets finding methods of ConnS, PatchS, and CPT. For CNNs in Table~\ref{tab:cnn}, the weight-level winning tickets is exit in different structures, VGG-9 and ResNet-19, in different datasets. And the spiking-based CNNs could outperform the normal CNNs under the extremely sparse case, which is pruning weights, as well as binarizing weights and activation simultaneously. For Transformer in Table~\ref{tab:vit}, the patch-level winning tickets are both exiting among normal and spiking-based Transformer in various datasets. Since the Spikeformer using convolutional operation in embedding, it makes its sparsity could be further improved by ConnS. 

\subsection{The Effect of Different Parameters}

\textbf{Pruning Rate:}
In our experiments, we try to pursue performance change when modifying the pruning rate for weight and patch-level winning tickets for different models under the event-based dataset. Therefore, for spiking parameters, we keep the Time Step $T=4$ and $\lambda=0.99$ for two models in various weight and activation conditions, like full-precision, binary and spiking. 
As presented in Figure~\ref{fig:3} (a), it illustrates the performance change for VGG-9 and TINY in DVS128Gesture. 
As the figures of VGG-9 show, normal, binary and spiking-based CNNs always go through an increase and decrease process. The weight-based winning tickets could attain the best performance when the pruning rate is set to $50\%$. Therefore, in the previous Table~\ref{tab:cnn}, we set the pruning rate to $50\%$ to compare with other dense performances. Additionally, in the figures of TINY, normal and spiking-based Transformers are sensitive to the pruning rate change. Both of them could tolerate a certain degree of patch pruning, whereas the performance would drop drastically when the pruning rate is beyond a threshold. It seems this threshold for different types of TINYs are all around $30\%$. In the Appendix, after going through a more detailed analysis, we ensure the best threshold is $31.4\%$ here. Additionally, we also ensure the performance when introducing ConnS into the process of finding patch-based winning tickets. When the pruning rate of ConnS is fixed at $48.7\%$ according to the analysis in the following subsection and Appendix, the concurrent weight and patch-based tickets of Spikeformer could display a similar change as others.



\noindent\textbf{Time Step}
In this paragraph, we conduct a thorough experiment to validate the effect of selecting a suitable Time Step for searching weight level and patch level winning tickets for VGG-9 and TINY in DVS128Gesture. In the meantime, for other parameters, the Decay Rate is $\lambda=0.99$, and the weight and patch level sparsity pruning rate is set to $50\%$ and $30\%$.
In Figure~\ref{fig:3} (b), as mentioned above, we select $T=1,2,4,6,8$ to verify the influence of Time Step on the performance of weight and patch level winning tickets; an obvious increase occurs in the sparse performance. 
From the theory of SNN, we could intuitively know the timestep could offer information gain following the increase of its value since $T$ times spiking simulation can be considered as multiple processing of the same input representation. This performance has already been verified in the dense case~\cite{fang2021deep,zheng2021going}. Nevertheless, through our experiments, even in the most extremely sparse case, this performance still exists no matter which kind of model structures are selected. Based on our concrete experimental results, increasing timestep from $T=1$ to $T=8$ with different pruning rates could achieve at most $+2.3\%$ and $+3.8\%$ increase in VGG-9 and TINY.


\noindent\textbf{Decay Rate:}
After thoroughly examining the relationship between Time Step $T$ and sparse performance, we also concentrate on another critical hyper-parameter Decay Rate $\lambda$ in SNNs. We select the $\lambda = 0.99, 0.8, 0.6, 0.4, 0.2$ as our variables to be validated. In the meantime, for other parameters, the Time Step is $T=4$, and the weight and patch level sparsity pruning rate is set to $50\%$ and $30\%$.
Our experiment results are presented in Fig.~\ref{fig:3} (c), modifying the decay rate could affect the final sparse accuracy among different model structures. All performances of various models also go through an increase and decrease process. Based on the illustration of the figure, change the timestep from $\lambda=0.99$ to $\lambda=0.2$ with different pruning rates could achieve at most $+2.1\%$ and $+1.3\%$ increase in VGG-9 and TINY.
However, according to our observation, the choice of $\lambda$ that inspires to produce the best sparse performance is related to the particular model size. The structure with a smaller model size could tolerate a more significant decay rate, but a larger model would suffer more damage when set with huge $\lambda$. 

\section{Connection and Patch Sparisty}


In this paragraph, we try to figure out the effect of different pruning rates of ConnS for fixed patch-level winning tickets in Transformers. For the parameters, the Time Step is $T=4$ and Decay Rate $\lambda=0.99$, and the patch level sparsity pruning rate is set to $30\%$. And the pruning rates of ConnS are adopted as $0, 0.2, 0.4, 0.6, 0.8$.
As presented in Figure~\ref{fig:4}, 
this paragraph identifies the phenomenon similar to the patch-level winning tickets in Transformers, where a certain threshold exists. When the sparsity level is below this threshold, it is possible to maintain the original performance of the neural network even with an increase in weight sparsity. However, exceeding this threshold significantly diminishes performance. The method of finding CPTs enables further enhancement of the sparsity level in spiking-based transformers, or Spikeformers, without compromising their original performance. This discovery opens up new avenues for optimizing the efficiency of Spikeformers while retaining their effectiveness.

\section{Conclusion}

This paper marks significant strides in the field of neural network optimization and efficiency. By extending the lottery tickets hypothesis to event-based datasets and spiking neural networks, it paves the way for more nuanced and efficient neural network architectures. The transition from traditional spike-based CNNs to advanced Spikeformers, coupled with the detailed analysis of SNN parameters, highlights the potential for SLTs in various levels and types of data. The proposed concurrent implementation of weight and patch-level SLTs in Spikeformers not only showcases the versatility of this approach but also sets a foundation for future research in neural network optimization. Finally, we develop a concurrent implementation method for Sparse Lottery Tickets at both weight and patch levels in Spikeformers represents a groundbreaking step in efficient network research.

\newpage
{
    \small
    \bibliographystyle{ieeenat_fullname}
    \bibliography{main}
}


\end{document}